\documentclass[10pt,twocolumn,letterpaper]{article}

\usepackage{cvpr}
\usepackage{times}
\usepackage{epsfig}
\usepackage{graphicx}
\usepackage{amsmath}
\usepackage{amssymb}
\usepackage{color}
\usepackage[utf8]{inputenc}

\usepackage{booktabs}
\usepackage{paralist} % for compactlist
\usepackage{subcaption}

\usepackage[bookmarks=false]{hyperref}

\cvprfinalcopy % *** Uncomment this line for the final submission

\def\cvprPaperID{7378} % *** Enter the CVPR Paper ID here

\DeclareMathOperator*{\argmin}{argmin}

\begin{document}
\title{Rethinking Zero-shot Video Classification: \\End-to-end Training for Realistic Applications}

\author{
Biagio Brattoli\thanks{Work done during an internship at Amazon.}\\
Heidelberg University\\
{\tt\small biagio.brattoli@iwr.uni-heidelberg.de}
\and
Joseph Tighe\\
Amazon\\
{\tt\small tighe@amazon.com}
\and
Fedor Zhdanov\\
Amazon\\
{\tt\small fedor@amazon.com}
\and
Pietro Perona\\
Amazon\\
{\tt\small perona@caltech.edu}
\and
Krzysztof Chalupka\\
Amazon\\
{\tt\small chalupkk@amazon.com}
}

\maketitle

%%%%%%%%% ABSTRACT
\begin{abstract}
Trained on large datasets, deep learning (DL) can accurately classify videos into hundreds of diverse classes. However, video data is expensive to annotate. Zero-shot learning (ZSL) proposes one solution to this problem. ZSL trains a model once, and generalizes to new tasks whose classes are not present in the training dataset. We propose the first end-to-end algorithm for ZSL in video classification. Our training procedure builds on insights from recent video classification literature and uses a trainable 3D CNN to learn the visual features. This is in contrast to previous video ZSL methods, which use pretrained feature extractors. We also extend the current benchmarking paradigm: Previous techniques aim to make the test task unknown at training time but fall short of this goal. We encourage domain shift across training and test data and disallow tailoring a ZSL model to a specific test dataset. We outperform the state-of-the-art by a wide margin. Our code, evaluation procedure and model weights are available at 
% \url{github.com/bbrattoli/ZeroShotVideoClassification}.
\textcolor{blue}{\href{https://github.com/bbrattoli/ZeroShotVideoClassification}{github.com/bbrattoli/ZeroShotVideoClassification}}.
\end{abstract}

%%%%%%%%% BODY TEXT
\section{Introduction}\label{sec:intro}

\begin{figure}
\centering
\includegraphics[width=.45\textwidth]{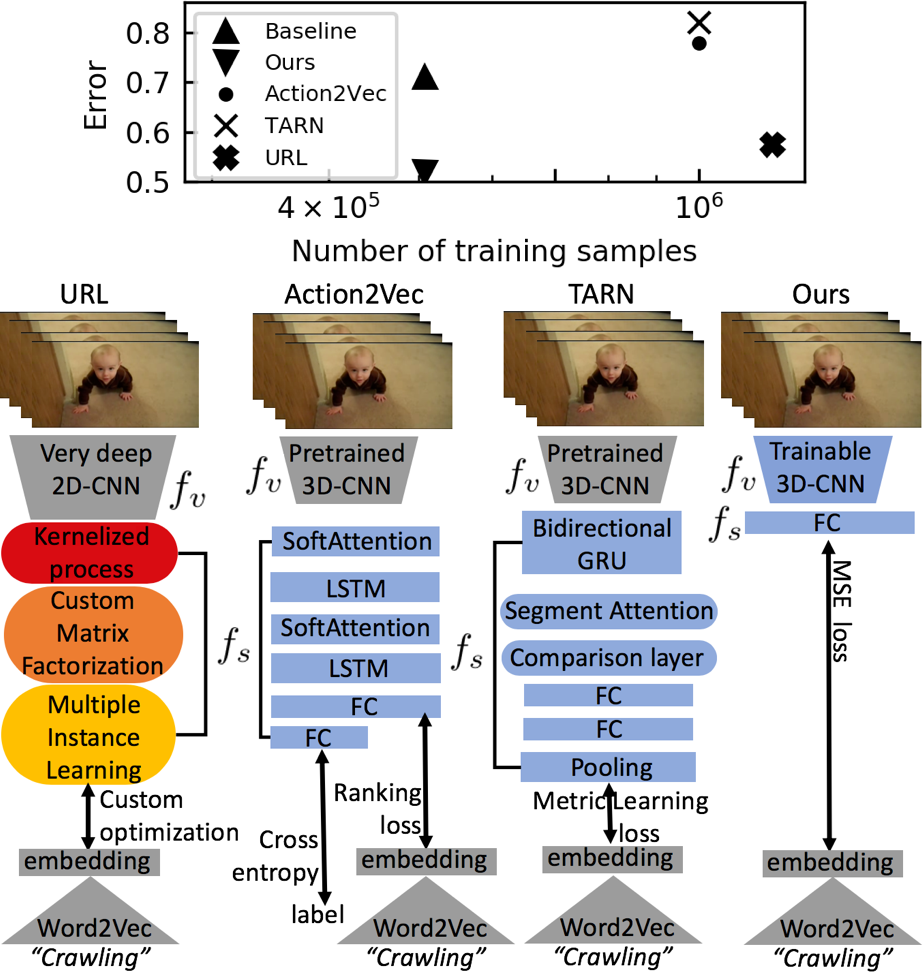}
\caption{(Top) Our model is state-of-the-art (error computed on the UCF test dataset.) (Bottom) Our e2e model is simple but powerful. URL~\cite{uar}, Action2Vec~\cite{action2vec} and TARN~\cite{tarn} are state-of-the-art approaches. Gray blocks represent modules fixed during training. Colors (blue, red, orange, yellow) indicate modules trained in separate stages.}
\label{fig:page1}
\end{figure}

Training image and video classification algorithms requires large training datasets~\cite{resnet,alexnet,c3d,r2plus1d,tsn}. With no task-specific training data available one may still attempt to train a model using related information and transfer the learned knowledge to classify previously unseen categories. This approach is called zero-shot learning (ZSL)~\cite{larochelle2008zero, palatucci2009zero} and it is quite successful in the image domain~\cite{mic, roth2020revisiting, DenseposeEvo20,dcesml,facenet,xian2017zero}.

We focus on ZSL for video action recognition, where data sourcing and annotation is particularly expensive. Since the set of possible human actions is huge, action recognition is a great ZSL testbed. Trained on large-scale academic datasets~\cite{activitynet,smtsmt,sports1m,kinetics,hmdb,ucf}, supervised 3D convolutional neural networks (CNNs) proved successful in this domain~\cite{slowfast,c3d,r2plus1d}. How well modern deep networks can recognize human actions in the ZSL setting is, however, an open question.

To our knowledge, all current ZSL methods for video recognition use pretrained visual embeddings~\cite{alexiou2016exploring, tarn,action2vec, mishra2018generative, piergiovanni2018learning, wang2017alternative, wang2017zero, xu2015semantic, xu2017transductive, xu2016multi, zhang2018cross, uar}. This provides a good tradeoff between training efficiency and using prior knowledge. Shallow trainable models then convert the pretrained representations to ZSL embeddings, as shown in Fig.~\ref{fig:page1}~(Bottom). Low training space complexity of shallow models allows them to benefit from long video sequences~\cite{c3d} and large feature extractors~\cite{resnet}. 

In contrast, state-of-the-art algorithms in the fundamental CV domains of image classification~\cite{resnet}, object detection~\cite{yolo,rcnn,sniper} and segmentation~\cite{deeplab,maskrcnn,pspnet} all rely on end-to-end (e2e) training. Representation learning is at the core of deep networks' success across machine learning domains~\cite{representation}, and deeper models can better utilize information available in large datasets~\cite{lots_of_data,resnet}. This poses a question: How can an e2e ZSL compete with current methods?

%Our contributions answer this question:
Our contributions involve multiple aspects of ZSL video classification:
\begin{compactenum}
  \item[\textbf{Novel Modeling:}] We propose the first e2e-trained model for zero-shot action recognition. The training procedure is inspired by modern supervised video classification practices. Fig.~\ref{fig:page1} shows that our method is simple, yet outperforms previous work. Moreover, we devise a novel easy pretraining technique that targets the ZSL scenario for video recognition.
  \item[\textbf{Evaluation Protocol:}] We propose a novel ZSL training and evaluation protocol that enforces a realistic ZSL setting. Extending the work of Roitberg~\etal~\cite{roitberg2018towards}, we test a single trained model on multiple test datasets, where sets of training and test classes are disjoint. In addition, we argue that training and test domains should not be identical.
  \item[\textbf{In-depth Analysis:}] We perform an in-depth analysis of the e2e model and a pretrained baseline. In a series of guided experiments we explore the characteristics of good ZSL datasets.
\end{compactenum}

Our model, training and evaluation code, are available at \textcolor{blue}{\href{https://github.com/bbrattoli/ZeroShotVideoClassification}{github.com/bbrattoli/ZeroShotVideoClassification}}.

\begin{figure}
\centering
\includegraphics[width=.45\textwidth]{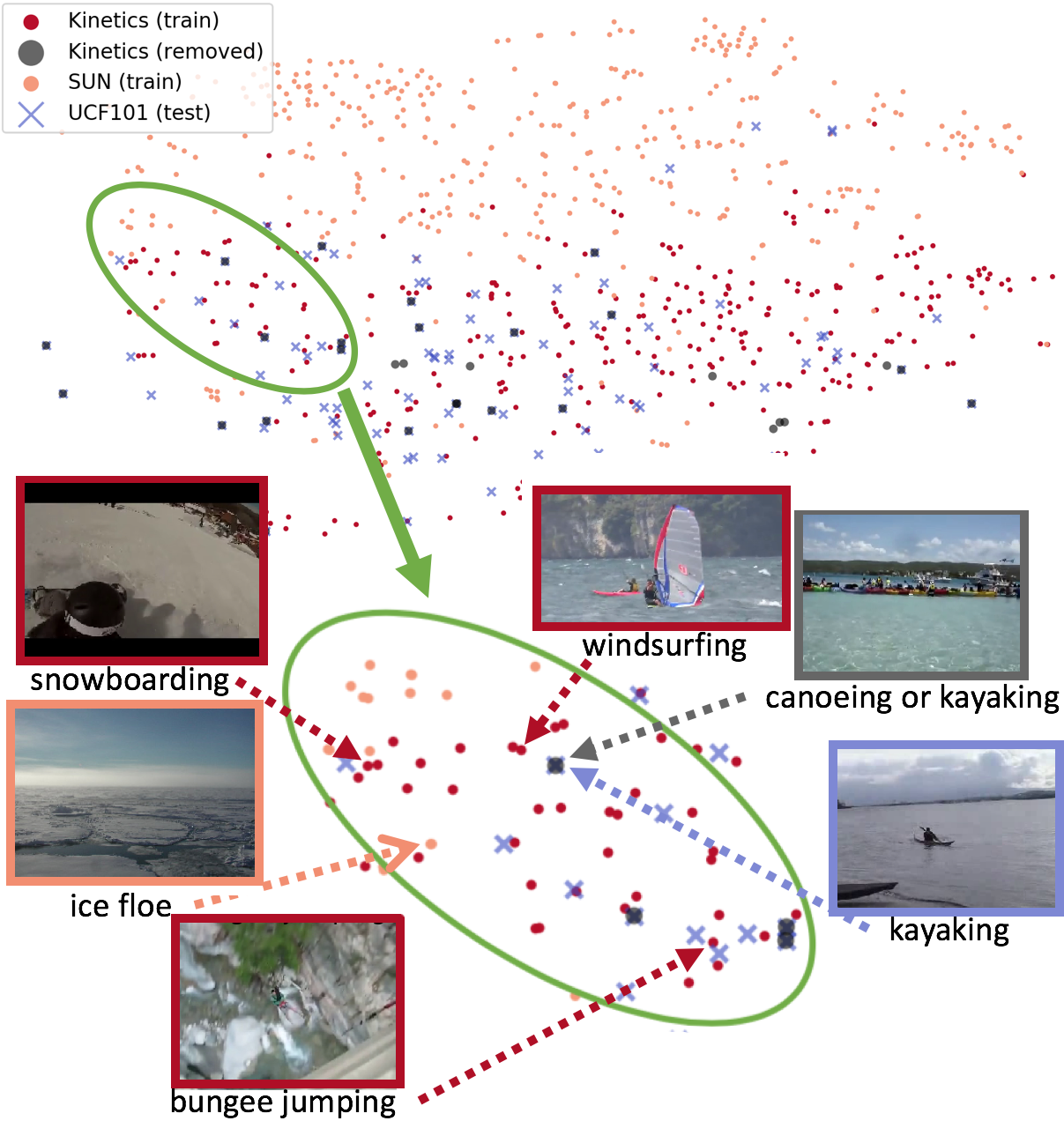}
\caption{Training and test classes, t-SNE~\cite{tsne} visualization of Word2Vec embeddings. Red dots represent training classes we used, and gray dots training classes we removed in order to separate training and test data. Crosses represent test classes. Pictures are actual dataset videoframes.}
\label{fig:tsne}
\end{figure}

\section{Related work}\label{sec:relatedwork}
We focus on \emph{inductive} ZSL in which test data is fully unknown at training time. There exists a body of literature on \emph{transductive} ZSL~\cite{alexiou2016exploring,mishra2018generative,wang2017alternative,wang2017zero,xu2017transductive,xu2015semantic,xu2016multi}, where test images or videos are available during training but test labels are not. We do not discuss the transductive approach in this work.

\textbf{Video classification:} Modern, DL-based video classification methods fall largely into two categories: 2D networks~\cite{twostream, tsn} that operate on 1-5 frame snippets and 3D networks~\cite{ourcvpr, oureccv, i3d, slowfast, r3d, action_finegrain, ourgcpr, c3d, r2plus1d} that operate on 16-128 frames. One of the earliest works of this type, Simonyan and Zisserman~\cite{twostream}, trained with only 1-5 frames sampled randomly from the video. At inference many more frames were sampled and the classifier outputs were averaged across all samples taken for a video clip. This implied that looking at a large chunk of the video was important during inference but wasn't strictly required during training. Wang~\etal~\cite{tsn} showed that sampling multiple frames throughout the video during training could improve performance, opening the question whether training also requires a large temporal context. However, a body of later work based on more powerful 3D networks~\cite{i3d, slowfast, c3d} showed that for most datasets sampling 16 frames during training is sufficient. Increasing training frame count from 16 to 128 improved performance only marginally.

In this work, we adapt the training-time sampling philosophy of state-of-the-art video classification to the ZSL setup. This allows us to train the visual embedding e2e. As a consequence, the overall architecture and inference procedure are very simple compared to previous work, and the results are state-of-the-art -- as shown in Fig.~\ref{fig:page1}.

\textbf{Zero shot video classification:}
The common practice in zero-shot video classification is to first extract visual features from video frames using a pretrained network such as C3D~\cite{c3d} or ResNet~\cite{resnet}, then trains a temporal model that maps the visual embedding to a semantic embedding space~\cite{tarn, chuang2016aaai, chuang2015aaai, chuang2016cvpr, chuang2016ijcv, action2vec, mettes2017, piergiovanni2018learning,zhang2018cross, uar}. Good generalization on semantic embeddings of class names means that the model can be applied to new videos where the possible output classes are not present in training data. Inference reduces to finding the test class whose embedding is the nearest-neighbor of the model's output. Word2Vec~\cite{word2vec} is commonly used to produce the ground-truth word embeddings. An alternative approach is to use manually crafted class attributes~\cite{idrees2017thumos}. We decided not to pursue the manual approach as it harder to apply in general scenarios.

\begin{figure}
\centering
\includegraphics[width=.45\textwidth]{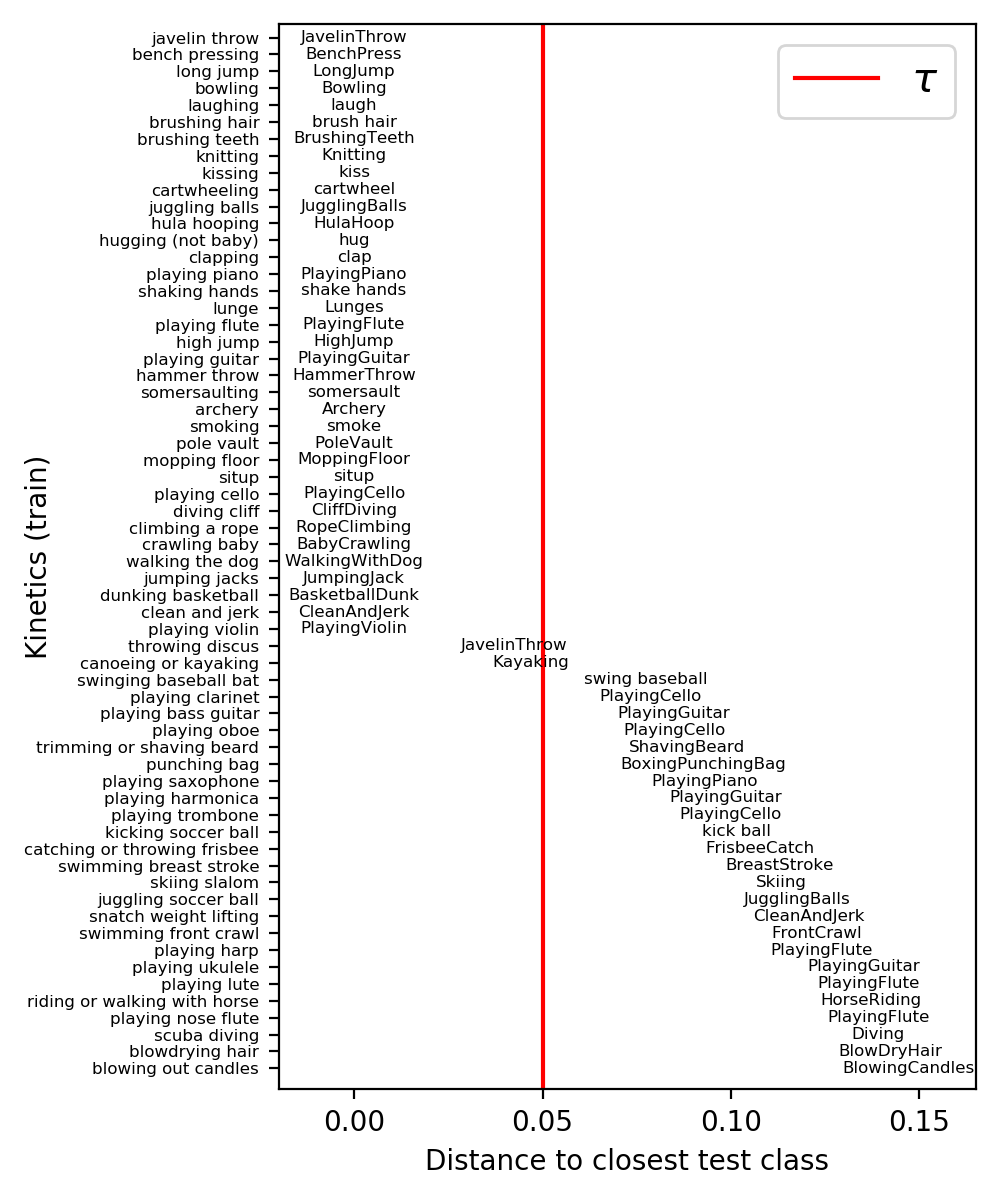}
\caption{Removing overlapping training and test classes. The y-axis shows Kinetics classes closest to the test sets UCF and HMDB. x-axis shows the distance (see Eq.~\ref{eq:dist}) of the corresponding closest test class. In our experiments, we removed training classes closer than $\tau=0.05$ to the test set -- to the left of the red line in the figure.}
\label{fig:overlapping}
\end{figure}

Two effective recent methods, Hahn~\etal~\cite{action2vec} and Bishay~\etal~\cite{tarn}, extract C3D features from 52 clips of 16 frames from each video. They then learn a recurrent neural network~\cite{gru, lstm} to encode the result as a single vector. Finally, a fully connected layer maps the encoded video into Word2Vec embedding. Fig.~\ref{fig:page1} illustrates this approach. Both ~\cite{action2vec} and \cite{tarn} use the same dataset for training and testing, after splitting the available dataset classes into two sets. Using a pretrained deep network is convenient because pre-extracted visual features easily fit in GPU memory, even for a large number of video frames. Alternative approaches use generative models to compensate for the gap between semantic and visual distributions~\cite{mishra2018generative, zhang2018visual}. Unfortunately, performance is limited by the inability to fine-tune the visual embedding. We show fine-tuning is crucial to generalize across datasets. 

Our work is similar to Zhu~\etal~\cite{uar} in that both methods learn a universal action representation that generalizes across datasets. However, their proposed model does not leverage the potential of 3D CNNs. Instead, they utilize the very deep ResNet200~\cite{resnet}, pretrained on ImageNet~\cite{imagenet1,imagenet2}, which cannot utilize temporal information. 

As pointed out by Roitberg~\etal~\cite{roitberg2018towards}, previous works train their models on actions overlapping with those of the the target dataset, violating ZSL assumptions. For example, Zhu~\etal~\cite{uar} train on the full ActivityNet~\cite{activitynet} dataset. This makes their results difficult to fairly compare with ours. Under our definition of ZSL (Sec.~\ref{sec:realistic}), Zhu~\etal have 23 classes in their training datasets that overlap with the test dataset. The situation is similar for all other methods to varying degrees.

\section{Zero-shot action classification}\label{sec:methods}
We first carefully define ZSL in the context of video classification. This will allow us to propose not only a new ZSL algorithm, but also a clear evaluation protocol that we hope will direct future research towards practical ZSL solutions. We stay within the inductive setting, as described in Sec.~\ref{sec:relatedwork}.

\subsection{Problem setting}\label{sec:problem}
A video classification task is defined by a training set (source) $D_s=\{(x_1, c_1), \cdots, (x_{N_s}, c_{N_s})\}$ consisting of pairs of videos $x$ and their class labels $c$, and a \mbox{video-label} test set $D_t$. In addition, previous work often uses pretraining datasets $D_p$ as explained in Sec.~\ref{sec:relatedwork}. 

Intuitively, ZSL is any procedure for training a classification model on $D_s$ (and possibly $D_p$) and then testing on $D_t$ \emph{where $D_t$ does not overlap with $D_s \cup D_p$}. %What "does not overlap" exactly means is a point of discussion. 
How this overlap is defined varies. Sec.~\ref{sec:realistic} proposes a definition that is more restrictive than those used by previous work, and forces the algorithms into a more realistic ZSL setting.

ZSL classifiers need to generalize to unseen test classes. One way to achieve this is using nearest-neighbor search in a semantic class embedding space.

Formally, given a video $x$, we infer the corresponding semantic embedding $z = g(x)$ and classify $x$ as the nearest-neighbor of $z$ in the set of embeddings of the test classes. Then, a trained classification model $M(\cdot)$ outputs
\begin{equation}
    M(x) = \argmin_{c\in D_t} \cos{\left(g(x), \text{W2V}(c)\right)}.
    \label{eq:classification}
\end{equation}
where $\cos$ is the cosine distance and the semantic embedding is computed using the Word2Vec function~\cite{word2vec} $\text{W2}V\colon\mathcal{C}\rightarrow\mathbb{R}^{300}$.

 The function $g = f_{s} \circ f_v$ is a composition of a visual encoder $f_v\colon x \mapsto y$ and a semantic encoder $f_s\colon y \mapsto z\in\mathbb{R}^{300}$.

\subsection{End-to-end training}\label{sec:e2e}
In previous work, the visual embedding function $f_v$ is either hand-crafted~\cite{xu2016multi, uar} or computed by a pretrained deep network~\cite{tarn, action2vec, wang2017zero, uar}. It is fixed during optimization, forcing model development to focus on improving $f_s$. Resulting models need to learn to transform fixed visual embeddings into meaningful semantic features and can be very complex, as shown in Fig.~\ref{fig:page1} (Bottom).
 
Instead, we propose to optimize both $f_v$ and $f_s$ at the same time. Such e2e training offers multiple advantages:
\begin{compactenum}
\item Since $f_v$ provides a complex computation engine, $f_s$ can be a simple linear layer (see Fig.~\ref{fig:page1}).
\item We can implement the full model using standard 3D CNNs. 
\item Pretraining the visual embedding on a classification task is not necessary. 
\end{compactenum}

End-to-end optimization using the full video is unfeasible due to GPU memory limitations. Our implementation is based on standard video classification methods which are effective even when only a small snippet is used during training, as discussed in detail in Sec~\ref{sec:relatedwork}. Formally, given a training video/class pair $(x, c) \in D_s$ we extract a snippet $x^t$ of $16$ frames at a random time $t \leq (\text{len}(x)-16)$. The network is optimized by minimizing the loss
\begin{equation}\label{eq:loss}
    L = \sum_{(x, c)\in D_s} \lVert \text{W2}V(c) - (f_s \circ f_v)(x^t)\rVert ^2.
\end{equation}
Inference procedure is similar but pools information from multiple snippets following Wang~\etal~\cite{tsn}. Sec.~\ref{sec:details} details both our training and inference procedures.

To better understand our method's performance under various experimental conditions, we implemented a baseline model that uses identical $f_s$, $f_v$ and training data, but fixes $f_v$'s weights to values pretrained on the classification task (available out-of-the-box in the most recent PyTorch implementation, see Sec.~\ref{sec:details}). This was necessary since we were not able to access implementations of any of the state-of-the-art methods~(\cite{tarn,action2vec,uar}). Unfortunately, our own re-implementations achieved results far below numbers reported by their authors, even with their assistance.

\subsection{Towards realistic ZSL}\label{sec:realistic}
To ensure that our ZSL setting is realistic, we extend the methods of~\cite{roitberg2018towards} that carefully separates training and test data. This is cumbersome to achieve in practice, and has not been attempted by most previous work. We hope that our clear formulation of the training and evaluation protocols will make it easy for future researchers to understand the performance of their models in true ZSL scenarios.

\textbf{Non-overlapping training and test classes:} Our first goal is to make sure that $D_s \cup D_p$ and $D_t$ have "non-overlapping classes". The simple solution -- to remove source class names from target classes or \emph{vice-versa} -- does not work, because two classes with slightly different names can easily refer to the same concept, as shown in Fig.~\ref{fig:overlapping}.
A distance between class names is needed.
Equipped with such a metric, we can make sure training and test classes are not too similar. 
Formally, let $d\colon\mathcal{C}\rightarrow\mathcal{C}$ denote a distance metric on the space of all possible class names $\mathcal{C}$, and let $\tau\in\mathbb{R}$ denote a similarity threshold. A video classification task fully respects the zero-shot constraint if
\begin{align}
    \forall c_s &\in D_s\cup D_p\textbf{, } \min_{c_t \in D_t} d(c_s, c_t) > \tau.
    \label{eq:nonoverlap}
\end{align}

A straightforward way to define $d$ is using semantic embeddings of class names. We define the distance between two classes to be simply 
\begin{align}
d(c_1, c_2) &= \cos({\text{W2V}(c_1), \text{W2V}(c_2)})
\label{eq:dist}
\end{align}
where $\cos$ indicates cosine distance. This is consistent with the use of the cosine distance in the ZSL setting as we do in Eq.~\ref{eq:classification}. 
Fig.~\ref{fig:tsne} shows an embedding of training and test classes after we removed from Kinetics classes overlapping with test data using the procedure outlined above. Fig.~\ref{fig:overlapping} shows the distribution of distances between training and test classes in our datasets. There is a cliff between distances very close to $0$ and larger than $0.1$. In our expeirments we use $\tau=0.05$ as a natural, unbiased threshold. 

\textbf{Different training and test video domains:} We argue that video domains of $D_s \cup D_p$ and $D_t$ should differ. In previous work, the standard evaluation protocol is to use one dataset for training and testing, using 10 random splits. This does not account for domain shifts that happen in real world scenarios due to data compression, camera artefacts, and so on. For this reason ZSL training and test datasets should ideally have disjoint video sources.

\begin{table}[t]
    \setlength{\tabcolsep}{2pt}
    \centering
    \begin{tabular}{llccc}  
    \toprule
    Dataset & VisualFeat & UCF & HMDB & Activity \\
    \toprule
    ObjEmb~\cite{mettes2017} & - & 40.4 & - & - \\
    URL~\cite{uar} & ResNet200 & 42.5 & 51.8 & - \\
    \midrule
    DataAug~\cite{xu2016multi} & - & 18.3 & 19.7 & -  \\
    InfDem~\cite{roitberg2018informed} & I3D & 17.8 & 21.3 & - \\
    Bidirectional~\cite{wang2017zero} & IDT & 21.4 & 18.9 & - \\
    FairZSL~\cite{roitberg2018towards} & - & - & 23.1 & - \\
    TARN~\cite{tarn} & C3D & 19 & 19.5 & - \\
    Action2Vec~\cite{action2vec} & C3D & 22.1 & 23.5 & - \\
    \midrule
    Ours(605classes) & C3D & 41.5 & 25.0 & 24.8 \\
    Ours(664classes) & C3D & 43.8 & 24.7 & - \\
    Ours(605classes) & R(2+1)D\_18 &44.1 & 29.8 & \textbf{26.6} \\
    Ours(664classes) & R(2+1)D\_18 & \textbf{48} & \textbf{32.7} & - \\
    \bottomrule
    \end{tabular}
    \caption{Comparison with the state-of-the-art on standard benchmarks. %We cite previous works' results as we were not able to reproduce their numbers. 
    We evaluate on half test classes following Evaluation Protocol 1 (Sec.~\ref{sec:eval_protocol}).
    Ours(605classes) indicates we removed all training classes that overlap with UCF, HMDB, or ActivityNet. Ours(664classes) indicates we removed only training classes overlapping with UCF and HMDB. We outperform previous work in both scenarios. Sec.~\ref{sec:relatedwork} argues that URL's results are not compatible with other works as their training and test sets overlap and their VisualFeat is an order of magnitude deeper. ObjEmb~\cite{mettes2017} uses a combination of FasterRCNN and GoogleNet.}
    \label{tab:sota}
\end{table}

\textbf{Multiple test datasets:} A single ZSL model should perform well on multiple test datasets. As outlined above, previous works train and test anew for each available dataset (typically UCF and HMDB). In our experiments, training happens only once on the Kinetics dataset~\cite{kinetics}, and testing on all of UCF~\cite{ucf}, HMDB~\cite{hmdb} and ActivityNet~\cite{activitynet}.

\subsection{Easy pretraining for video ZSL}\label{sec:sun}
In a real-world scenario a model is trained once and then deployed on diverse unseen test datasets. A large and diverse training dataset is crucial to achieve good performance. Ideally, the training dataset would be tailored to the general domain of inference -- for example, a strong ZSL surveillance model to be deployed at multiple unknown locations would require a large surveillance and action recognition dataset.

Sourcing and labeling domain-specific video datasets is, however, very expensive. On the other hand, annotating images is considerably faster. Therefore, we designed a simple dataset augmentation scheme which creates synthetic training videos from still images. Sec.~\ref{sec:experiment} shows that pretraining our model using this dataset boosts performance, especially if available training data is small. 

We convert images to videos using the Ken Burns effect: a sequence of crops moving around the image simulates video-like motion. Sec.~\ref{sec:datasets} provides more details.

Our experiments focus on the action recognition domain. In action recognition (as well as in many other classification tasks), location and scenery of the video is strongly predictive of action category. Because of this we choose SUN~\cite{sun}, a standard scene recognition dataset. %, to a video dataset using the procedure outlined above. 
Fig.~\ref{fig:tsne} shows the complete class embedding of our the scene dataset's class names.

\section{Experimental setup}\label{sec:implementation}
To facilitate reproducibility, we describe our training and evaluation protocols in detail. The protocols propose one way of training and evaluating ZSL models that is consistent with our definitions in Sec.~\ref{sec:realistic}. 

\subsection{Datasets}
\label{sec:datasets}
%\kcomment{move this to Sec 4}
\textit{UCF101}~\cite{ucf} has 101 action classes primarily focused around sports, with 13320 videos sourced from YouTube.  
\textit{HMDB51}~\cite{hmdb} is divided into 51 human actions focused around sports and daily activities and contains 6767 videos sourced from commercial videos and YouTube.
\textit{ActivityNet}~\cite{activitynet} contains 27,801 untrimmed videos divided in 200 classes focusing on daily activities with videos sourced using web search. We extracted only the labeled frames from each video.
\textit{Kinetics}~\cite{kinetics} is the largest currently available action recognition dataset, covering a wide range of human activity. The first version of the dataset contains over 200K videos divided in 400 categories. The newest version has 700 classes for a total of 541624 videos sourced from YouTube.
\textit{SUN397}~\cite{sun} (see Sec.~\ref{sec:sun}) is a scene understanding image dataset. It contains 397 scene categories for a total of over 100K high-resolution images. We converted it to a simulated video dataset using the Ken Burns effect: To create a 16-frame video from an image, we randomly choose "start" and "end" crop locations (and crop sizes) in the image, and linearly interpolate to obtain 16 crops. Each of them are then resized to $112\times112$.

\begin{table}[t]
    \setlength{\tabcolsep}{3pt}
    \centering
    \begin{tabular}{lcccccc}  
    \toprule
    Method & \multicolumn{2}{c}{UCF}& \multicolumn{2}{c}{HMDB} & \multicolumn{2}{c}{Activity} \\
    \cmidrule(r{4pt}){2-3} \cmidrule(r{4pt}){4-5} \cmidrule(r{4pt}){6-7}
     & Top-1 & Top-5 & Top-1 & Top-5 & Top-1 & Top-5 \\
    \midrule
    ObjEmb~\cite{mettes2017} & 32.8 & - & - & - & - & - \\
    URL~\cite{uar} & 34.2 & - & - & - & - & - \\
    664classes & \textbf{37.6} & \textbf{62.5} & \textbf{26.9} & \textbf{49.8} & - & - \\
    605classes & 35.3 & 60.6 & 24.8 & 44.0 & \textbf{20.0} & \textbf{42.7} \\
    \bottomrule
    \end{tabular}
    \caption{Evaluation on all test classes. In contrast to Table~\ref{tab:sota}, here we report results of our method applied to all three test datasets using Evaluation Protocol 2 (Sec.~\ref{sec:eval_protocol}). We applied a single model trained on classes dissimilar from all of UCF, HMDB and ActivityNet. Nevertheless, we outperform URL~\cite{uar} on UCF101. ObjEmb and URL authors do not report results on full HMDB51. Remaining previous work do not report results on neither full UCF101 nor full HMDB51.}
    \label{tab:allclasses}
\end{table}

\subsection{Training protocol}
\label{sec:training_protocol}
Our experiments in Sec.~\ref{sec:experiment} use two training methods:

% \begin{compactenum}
% \item 

\textbf{Training Protocol 1:} Remove from Kinetics 700 all the classes whose distance to any class in $\text{UCF} \cup \text{HMDB}$ is smaller than $\tau$ (see Eq.~\ref{eq:dist}). This results in a subset of Kinetics with 664 classes, which we call Kinetics 664. As explained in Sec.~\ref{sec:realistic}, this setting is already more restrictive than that of the previous methods, which train new models for each test dataset.

\textbf{Training Protocol 2:} Remove from Kinetics 700 all the classes whose distance to any class in $\text{UCF} \cup \text{HMDB }\cup ActivityNet$ is smaller than $\tau$ (see Eq.~\ref{eq:dist}). This results in a subset of Kinetics with 605 classes which we call Kinetics 605. This setting is even more restrictive, but is closer to true ZSL. Our goal is to show that it is possible to train a single ZSL model that applies to multiple diverse test datasets.
% \end{compactenum}

Figure~\ref{fig:tsne} shows a t-SNE projection of the semantic embeddings of all Kinetics 700 classes, as well as the 101 UCF classes and the classes we removed to obtain Kinetics 664. 

\subsection{Evaluation protocol}
\label{sec:eval_protocol}
We tested our model using two protocols: the first follows Sec.~\ref{sec:realistic} to emulate a true ZSL setting, the second is compatible with previous work. Both Evaluation Protocols apply the same model to multiple test datasets.

\textbf{Evaluation Protocol 1:} 
In order to make our results comparable with previous work, we use the following procedure: Randomly choose half of the test dataset's classes, 50 for UCF and 25 for HMDB. Evaluate the classifier on that test set. Repeat ten times and average the results for each test dataset.

\textbf{Evaluation Protocol 2:} Previous work uses random training/test splits of UCF~\cite{ucf} and HMDB~\cite{hmdb} to evaluate their algorithms. However, we train on a separate dataset Kinetics 664 / 605 and can test on full UCF and HMDB. This allows us to return more realistic accuracy scores. The evaluation protocol is simple: evaluate the classifier on all 101 UCF classes and all 51 HMDB classes.

\subsection{Implementation details}
\label{sec:details}
In our experiments, $f_v$ (see Sec.~\ref{sec:problem}) is the PyTorch implementation of R(2+1)D\_18~\cite{r2plus1d} or C3D\cite{c3d}. In the pretrained setting, we use the out-of-the-box R(2+1)D\_18 pretrained on Kinetics 400\cite{kinetics}, while C3D is pretrained on Sports-1M\cite{sports1m}. In the e2e setting, we initialize the model with the pretrained=False argument. 
The visual embedding $f_v(x)$ is BxTx512 where B is the batch size and T is the number of clips per video. We use $T=1$ for training, and $T=25$ for evaluation in Tables~\ref{tab:sota}~and~\ref{tab:allclasses}. The clips are 16 frames long and we choose them following standard protocols established by Wang~\etal~\cite{tsn}. We average $f_v(x)$ across time (video snippets) similarly to previous approaches~\cite{c3d, uar}. $f_{s}$ is a linear classifier with 512x300 weights. The output of $f_{s} \circ f_v$ is of shape Bx300.

We follow standard protocol in computing semantic embeddings of class names~\cite{tarn, xu2015semantic, uar}. Word2Vec~\cite{word2vec} -- in particular, the gesim~\cite{gesim} Python implementation -- encodes each word. We average multi-word class names. In rare cases of words not available in the pretrained W2V model (for example, 'rubiks' or 'photobombing') we manually change the words (see the code for more details). %Our code describes this process in detail. 
Formally, for a class name consisting of N words $c=[c^1, \cdots, c^N]$, we embed it as $\text{W2}V(c) = \sum_{i=1}^N \text{W2}V(c^i) \in \mathbb{R}^{300}$. We set $\tau$ to $0.05$ following the analysis in Sec.~\ref{sec:realistic} based on Fig.~\ref{fig:overlapping}.%following Roitberg~\etal~\cite{roitberg2018towards}.

To minimize the loss of Eq.~\ref{eq:loss} we use the Adam optimizer~\cite{adam}, starting with a learning rate of $1e-3$. Batch size is 22 snippets, with 16 frames each. The model trained for 150 epochs, with a tenfold learning rate decrease at epochs 60 and 120. All experiments are performed on the Nvidia Tesla V100 GPU. 

Following~\cite{c3d}, we reshaped each frame's shortest side to 128 pixels, and cropped a random 112x112 patch on training and the center patch on inference. %The random crop is fixed for each frame of the same snippet.

\begin{figure}
\centering
\includegraphics[width=.45\textwidth]{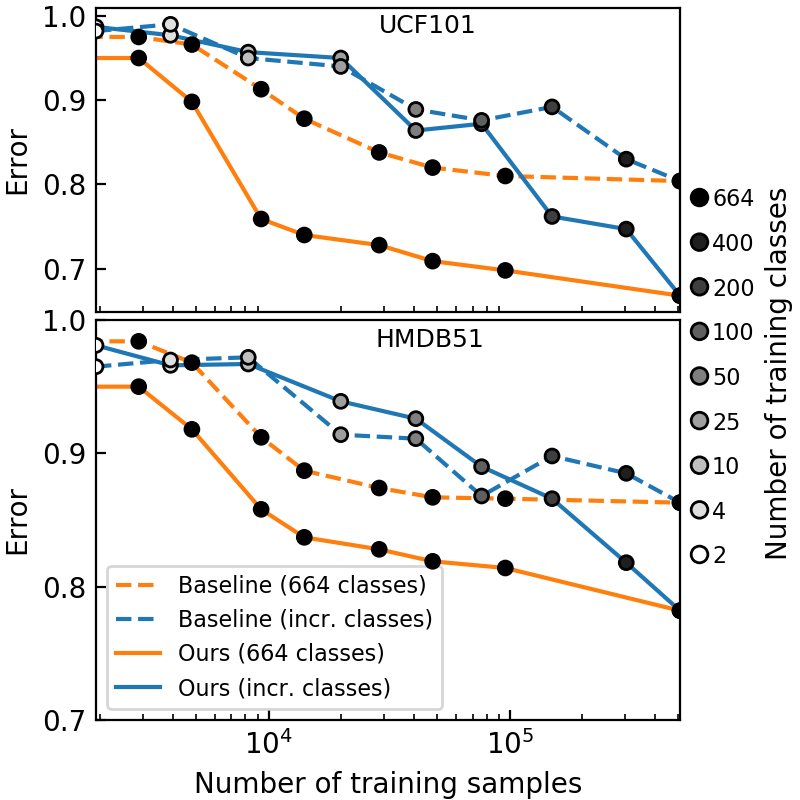}
\caption{Number of training classes matters in ZSL. Orange curves show performance on subsets of Kinetics 664, as we keep all the training classes and increase the subset size. The blue curves, whose markers become progressively brighter, indicate a separate experiment where we increased the number of training classes starting from 2, all the way up to 664 (Sec.~\ref{sec:comparebaseline}). For any given training dataset size, performance on test data is much better with more training classes. In addition, when few training classes are available the e2e model is not able to outperform the baseline.}
\label{fig:classes}
\end{figure}

\begin{figure}
\centering
\includegraphics[width=.40\textwidth]{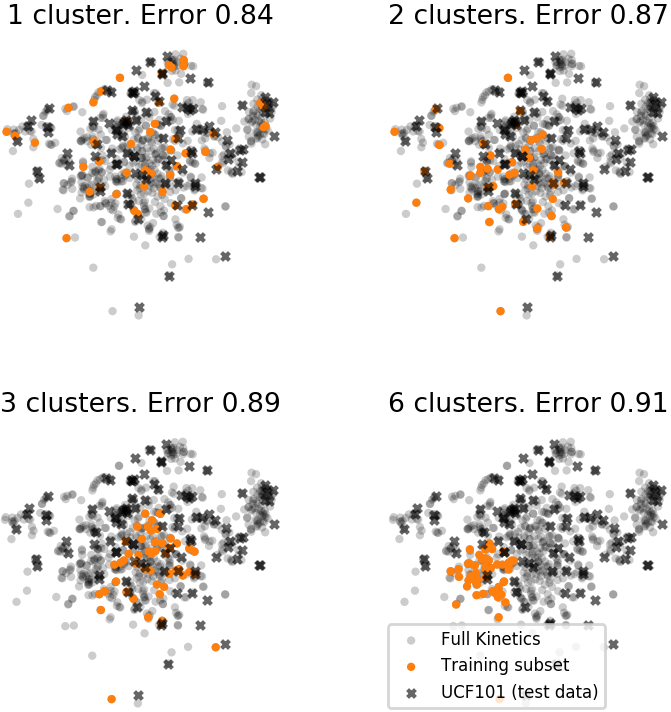}
\caption{Diverse training classes are good for ZSL. Here we trained our algorithm on subsets of 50 Kinetics 664 classes. (Top left) Training classes picked uniformly at random. (Top right) We clustered Word2Vec embeddings of classes into two clusters, then trained and evaluated separately using each cluster, and averaged the results. (Bottom) Here we averaged the results of training using three and six clusters. The figure shows that the more clusters, the less diverse the training classes were semantically. At the same time, less diversity caused higher errors.}
\label{fig:cluster}
\end{figure}

\begin{figure}
\centering
\includegraphics[width=.45\textwidth]{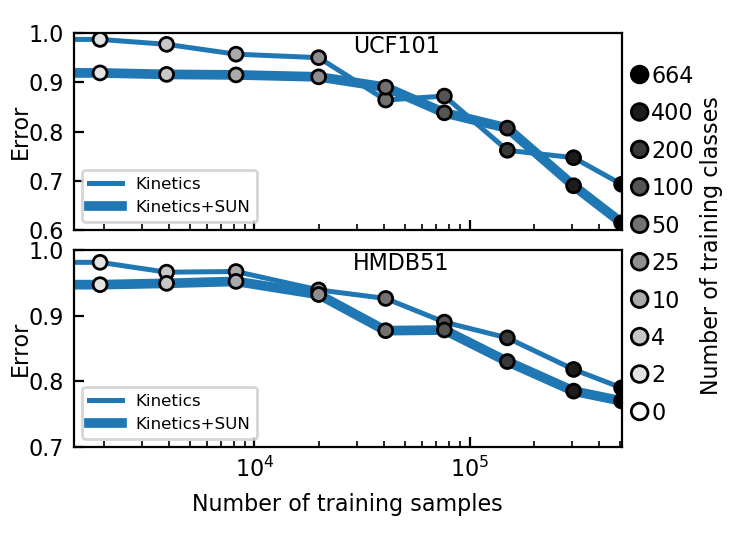}
\caption{Augmented pretraining with videos-from-images. We trained our algorithm on progressively smaller subsets of Kinetics 664 classes (Sec.~\ref{sec:comparebaseline}). We compared the results to training on the same dataset, after pretraining the model on our synthetic SUN video dataset (Sec.~\ref{sec:expsun}). The pretraining procedure boosts performance up to 10\% points.}
\label{fig:sun}
\end{figure}

\section{Results}\label{sec:experiment}
Our experiments have two goals: compare our method to previous work and investigate our method's performance vs the baseline (see Sec.~\ref{sec:e2e}.) The first is necessary to validate that e2e ZSL on videos can outperform more complex approaches that use pretrained features. The latter will allow us to understand under what conditions e2e training can be particularly beneficial. 

\subsection{Comparison to the state of the art}
Table~\ref{tab:sota} compares our method to existing approaches. We followed our Training and Evaluation Protocol 1, as described in Sections~\ref{sec:training_protocol}~and~\ref{sec:eval_protocol}. Our protocols are more restrictive than that of previous methods: we removed training classes that overlap with test classes, introduced domain shift, and applied one model to multiple test datasets. Despite this, we outperform previous video-based methods by a large margin. Furthermore, when testing on UCF we outperform URL~\cite{uar} which uses a network an order of magnitude deeper than ours -- 18 vs 200 layers -- and 23 classes overlap between training and testing (see Sec.~\ref{sec:relatedwork}).
%uses a very complex semantic embedding function $f_s$ (see Fig.~\ref{fig:page1}).

\subsection{Comparison to a baseline method}
\label{sec:comparebaseline}
Our baseline method described in Sec.~\ref{sec:e2e} uses a fixed, pretrained visual feature extractor but is otherwise identical to our e2e method. This allows us to study the benefits of e2e training under Evaluation Protocol 2, (see Sections~\ref{sec:training_protocol}~and~\ref{sec:eval_protocol}). Using all test classes provides a more direct evaluaition of the method.
%This paints a clearer picture of the algorithms' performance, as we use all classes of UCF and HMDB on test time.

\textbf{Training dataset size:}
To investigate the effect of training set size on performance we subsampled Kinetics 664 uniformly at random, then re-trained and re-evaluated the model. %both our algorithm and the baseline. 
Fig.~\ref{fig:classes} shows that the e2e algorithm consistently outperforms the baseline on both datasets. Both algorithms' performance is worse with smaller training data. However, the baseline flattens out at about 100K training datapoints, whereas our method's error keeps decreasing. This is expected, as the e2e model has more capacity.

\textbf{Number of training classes:}
In many video domains diverse data is difficult to obtain. Small datasets might not only have few datapoints, but also contain only a few training classes. We show that the number of training classes can impact ZSL results as much as training dataset size.

To obtain Fig.~\ref{fig:classes} we subsampled Kinetics 664 class-wise. We first picked 2 Kinetics 664 classes at random, and trained the algorithm on those classes only. We repeated the procedure using 4, 10, 25, 50, 100, 200, 400 and all 664 classes. Naturally, the fewer classes the fewer datapoints the training set contained. This results are compared in Fig.~\ref{fig:classes} with the procedure described above, where we removed Kinetics datapoints at random -- independent of their classes. 

The figure shows that it is better to have few training samples from a large number of classes rather than many from a very small number of classes.
This effect is more pronounced for the e2e model rather than the baseline.

\textbf{Training dataset class diversity:}
We showed that ZSL works better with more training classes. If we have a limited budget for collecting classes and datapoints, how should we choose them? We investigated whether the set of training classes should emphasize fine differences (e.g. "shooting basketball" vs "passing basketball" vs "shooting soccerball" and so on) or diversity. 

In Fig.~\ref{fig:cluster} we selected $50$ training classes in four ways: (Top Left) We randomly choose 50 classes from the whole Kinetics 664 dataset, trained the algorithm on these classes, and ran inference on the test set. We repeated this process ten times and averaged inference error. (Top Right) We clustered the 664 classes into 2 clusters in the Word2Vec embedding space, and chose 50 classes at random within one of the clusters, trained and ran inference. We then repeated the procedure ten times and averaged the result. (Bottom) Here we chose 50 classes in one of 3 clusters (Left) and one of 6 clusters (Right), trained, and averaged inference results of 10 runs. The figure shows that test error for our method increases as class diversity decreases. This result is not obvious, since the task becomes harder with increasing class diversity.

\subsection{Easy pretraining with images}\label{sec:expsun}
\begin{figure}
\centering
\includegraphics[width=.42\textwidth]{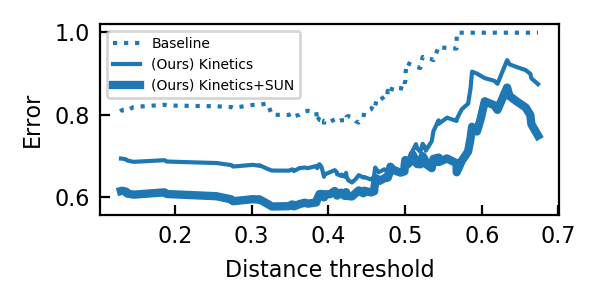}
\caption{Error as test classes move away from training. For each UCF101 test class, we computed its distance to 10 nearest neighbors in the training dataset. We arranged all such distance thresholds on the x-axis. For each threshold, we computed the accuracy of the algorithms \emph{on test classes whose distance from training data is larger than the threshold}. In other words, as x-axis moves to the right,
the model is evaluated on cumulatively smaller, but harder test sets.}
\label{fig:error_distance}
\end{figure}

Previous section showed that class count and diversity are important drivers of ZSL performance. This inspired us to develop the pretraining method described in Sec.~\ref{sec:sun}: we pretrain our model on a synthetic video dataset created from still images from the SUN dataset. Fig.~\ref{fig:sun} shows that  this simple procedure consistently decreases test errors by up to 10\%. In addition, Fig.~\ref{fig:error_distance} shows that this initialization scheme makes the model more robust to large domain shift between train and test classes. The following section describes the latter finding in more detail.

\subsection{Generalization and domain shift}
A good ZSL model generalizes well to classes that differ significantly from training classes. To investigate the performance of our models under heavy domain shift, we computed the accuracy on subsets of test data with a growing distance from the training dataset. We first trained our model on Kinetics 664. Then, for a given distance threshold $\tau$ (see Sec.~\ref{sec:realistic}), we computed accuracy on the set of UCF classes whose mean distance from the closest 10 Kinetics 664 classes is larger than $\tau$. Fig.~\ref{fig:error_distance} shows that the baseline model's %(pretrained on a large dataset but 
(not trained e2e) performance drops to zero at around $\tau\sim0.57$. Our method performs much better, never dropping to zero accuracy for high thresholds. Finally, using the SUN pretraining further increases performance.

\subsection{Ablation study}
Table~\ref{tab:ablation} studies contributions of different elements of our model to its performance. The performance is low when the visual embedding is fixed. The e2e approach improves the performance by a large margin. Our class augmentation method further boosts performance. Finally it helps to extract linearly spaced snippets from a video on testing, and average their visual embeddings. Using $25$ snippets improves considerably the performances without influencing the training time of the model.

\begin{table}[t]
    \setlength{\tabcolsep}{3pt}
    \centering
    \begin{tabular}{ccccccc}  
    \toprule
    \multicolumn{3}{c}{UCF101 accuracy} & \multicolumn{2}{c}{50 classes} & \multicolumn{2}{c}{101 classes} \\
    \cmidrule(r{4pt}){4-5} \cmidrule(r{4pt}){6-7}
    e2e & Augment & Multi & Top-1 & Top-5 & Top-1 & Top-5 \\
    \midrule
     & &  & 26.8 & 55.5 & 19.8 & 40.5\\
    \checkmark & &  & 43.0 & 68.2 & 35.1 & 56.4\\
    \checkmark & \checkmark &  & 45.6 & 73.1 & 36.8 & 61.7\\
    \checkmark &  & \checkmark & 48.0 & 74.2 & 37.6 & 62.5 \\
    \checkmark & \checkmark & \checkmark & \textbf{49.2} & \textbf{77.0} & \textbf{39.8} & \textbf{65.6} \\
    \bottomrule
    \end{tabular}
    \caption{Ablation study. Numbers represent classification accuracy. ``50 classes'' uses Evaluation Protocol 1 (Sec.~\ref{sec:eval_protocol}.) ``101 classes'' uses Evaluation Protocol 2. e2e: training the visual embedding as opposed to fixed, pretrained baseline (Sec.~\ref{sec:e2e}). Augment: pretrain using the SUN augmentation scheme (Sec.~\ref{sec:expsun}). Multi: At test time, extract multiple snippets from each video and average the visual embeddings (Sec.~\ref{sec:details}).}
    \label{tab:ablation}
\end{table}

\section{Conclusion}
\label{sec:conclusion}
We followed practices from recent video classification literature to train the first e2e system for video recognition ZSL. Our evaluation protocol is stricter than that of existing work, and measures more realistic zero-shot classification accuracy. Even under this stricter protocol, our method outperforms previous works whose performance was measured with training and test sets overlapping and sharing domains. Through a series of directed experiments, we showed that a good ZSL dataset should have many diverse classes. Guided by this insight, we formulated a simple pretraining technique that boosts ZSL performance.

Our model is easy to understand and extend. 
Our training and evaluation protocols are easy to use with alternative approaches. 
We made our code available at \textcolor{blue}{\href{https://github.com/bbrattoli/ZeroShotVideoClassification}{github.com/bbrattoli/ZeroShotVideoClassification}} to encourage the community to build on our insights and create a strong foundation for future video ZSL research.\\

\textbf{Acknowledgement.} 
We thank Amazon for generously supporting the project, and Alina Roitberg for a productive discussion on the evaluation protocol.
%We thank Amazon for providing the computational means and Alina Roitberg for a productive discussion about the evaluation protocol.

\clearpage
{\small
\bibliographystyle{ieee_fullname}
\bibliography{egbib}
}

\end{document}

%%%%%%%%% TITLE
\title{SUPPLEMENTARY MATERIAL\\ Rethinking Zero-shot Video Classification: \\End-to-end Training for Realistic Applications}

% \and

\maketitle
%\thispagestyle{empty}

%%%%%%%%% SUPPLEMENTARY
\section{Backbone choice}
Supplementary Table~\ref{tab:architectures} compares the accuracy of three 3D convolutional backbones on two kinetics versions using our Training Protocol 1 (Sec.~4.2, Main Text). %Evaluation followed Evaluation Protocol 2. 
For this comparison we also tried using the full Kinetics 400/700 datasets, without removing overlapping test classes. The table shows that adding the 6\% of the training classes most overlapping with the test set yields an unexpected $>$40\% accuracy boost for UCF and 25\% on HMDB. This proves that the zero-shot learning constraint is non-trivial.

\begin{table}[t]
    \setlength{\tabcolsep}{4pt}
    \centering
    \begin{tabular}{lllcccc}
    \toprule
    &&& %UCF101 & HMDB51 \\
    \multicolumn{2}{c}{UCF}& \multicolumn{2}{c}{HMDB}  \\
    \cmidrule(r{4pt}){4-5} \cmidrule(r{4pt}){6-7}
    Network & Train & classes & 50 & 101 & 25 & 51 \\
    % \midrule
    % URL[54] & - & - & 42.5 & 34.2 & - & - \\
    \midrule
    C3D & K400 & 361 & 33.7 & 25.7 & 17.0 & 13.3 \\
    R3D\_18 & K400 & 361 & 37.2 & 29.0 & 20.4 & 16.8\\
    R(2+1)D\_18 & K400 & 361 & 38.7 & 30.6 & 22.0 &18.1 \\
    \midrule
    C3D & K700 & 664 & 40.3 & 33.1 & 22 & 17.0 \\
    R3D\_18 & K700 & 664 & 41.2 & 34.2 & 23.6 & 19.0 \\
    R(2+1)D\_18 & K700 & 664 & \textbf{43.0} &\textbf{35.0}& \textbf{25.8} & \textbf{20.6}\\
    \midrule
    R(2+1)D\_18 & K400 & 400 & 50.1 & 44.5 & 27.2 & 22.5 \\
    R(2+1)D\_18 & K700 & 700 & 54.6 & 49.7 & 30.5 & 25.6 \\
    \bottomrule
    \end{tabular}
    \caption{Accuracy of different backbone architectures trained on the first (K400) and last (K700) version of Kinetics~[19]. The models are evaluated %following Evaluation Protocol 2 
    on a single clip (16 frames).}
    \label{tab:architectures}
\end{table}

\begin{figure}
\centering
\includegraphics[width=.50\textwidth]{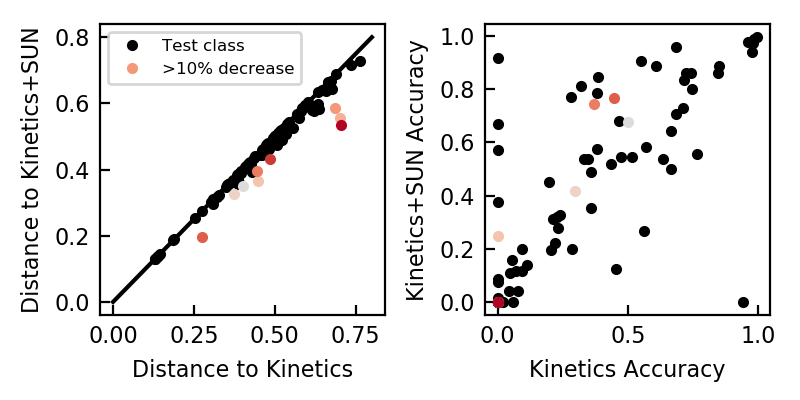}
\caption{Each dot represents a UCF101 test class. Test class accuracy (right) and distance (left) to the train set (Kinetics664) for two models: one with random initialization and one pretrained on SUN (see Sec~3.4, Main Text). A colored dot indicates a test class that reduces its distance to the train set by more than 10\% when SUN is included on training.}
\label{fig:accuracy_distance_ratio}
\end{figure}
%\begin{figure}
%\centering
%\includegraphics[width=.50\textwidth]{images/accuracy_distance_noratio.png}
%\caption{Each dot represents a UCF101 test class. \textbf{Top:} Test class accuracy (right) and distance (left) to the train set (Kinetics664) for two models: one with random initialization and one pretrained on SUN (see Sec~3.4, Main Text). A colored dot indicates a test class that reduces its distance to the train set by more than 10\% when SUN is included on training. \textbf{Bottom:} combines distance and accuracy plots using the ratio between the two models. The result shows that there is no correlation between train-test domain distance and performances. This indicates that the improvement is due to a better visual representation and not because of the problem becoming easier with the addition of SUN.}
%\label{fig:accuracy_distance_ratio}
%\end{figure}

\section{SUN pretraining: easier task or better representation?}
Section~3.4 (Main Text) shows that pretraining on a scenes dataset (SUN397) improves ZSL performance. In this section, we ask whether the boost is due to better model generalization or simply because the source domain becomes closer to the target domain. 

Per each UCF101 test class, Sup.~Fig.~\ref{fig:accuracy_distance_ratio} shows the W2V distance to Kinetics train classes as well as (Kinetics + SUN) train classes. Test classes that got more than 10\% closer to training data are marked in color. The right subplot, however, shows that the model trained on (Kinetics + SUN) boosts the accuracy of many classes -- in particular, the accuracy of many classes that are \emph{not} among the colored ones rose significantly. The model pretrained on SUN data increases performance on many classes which are not close to SUN data. We conclude that pretraining on SUN allows the model to generalize better over almost all test classes, not only the ones close to SUN data.

\section{Training class diversity}
\begin{figure}
\centering
\includegraphics[width=.45\textwidth]{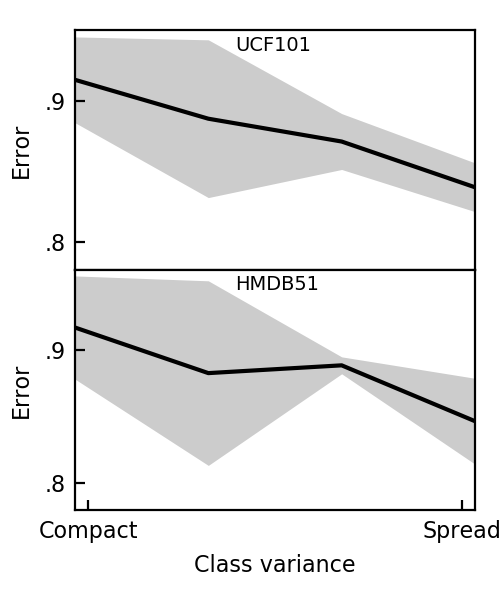}
\caption{Performance of the e2e model trained on 50 Kinetic664 classes and tested on UCF and HMDB. The 50 classes are chosen based on diversity in their W2V embedding (see Fig.~5, Main Text, for details). The more semantically diverse the training classes, the lower the error.}
\label{fig:variance}
\end{figure}

We expand the analysis of Sec.~5.2 and Fig.~5, Main Text, by testing the influence of training class diversity on both UCF and HMDB. Sup.~Fig.~\ref{fig:variance} correlates model performance with training class density. For this experiment, we selected 50 train classes with different density in the Word2Vec space, using the same clustering approach we used in Sec.~5.2. Per each diversity value, we select 50 classes and train a model multiple times to compute the standard deviation. Sup.~Fig.~\ref{fig:variance} shows that test error decreases as training classes become more diverse. At the same time, the standard deviation decreases, indicating that for compact classes, the performance highly depends on where in the class space we sample the classes, which is something we only know once the test set is available.

This outcome is not obvious, since we might expect the task to become harder when class variance increases (given the same number of training datapoints). However, we do not observe decrease in performance. Therefore, we can conclude that the model can only benefit from a high variety within the train class distribution. This new insight can be useful during training dataset collection.

\section{Analyze the model capability action per action}
What does better or worse accuracy indicate for specific classes? We break down the change in performance between models for each UCF101 test class. 

\subsection{Direct comparison by sorting classes}
In Sec.~5, Main Text, we evaluated the model using error aggregated over all the test classes. It is also interesting to know whether the network is getting better at recognizing specific classes, or improves across the board?

\begin{figure}
\centering
\includegraphics[width=.49\textwidth]{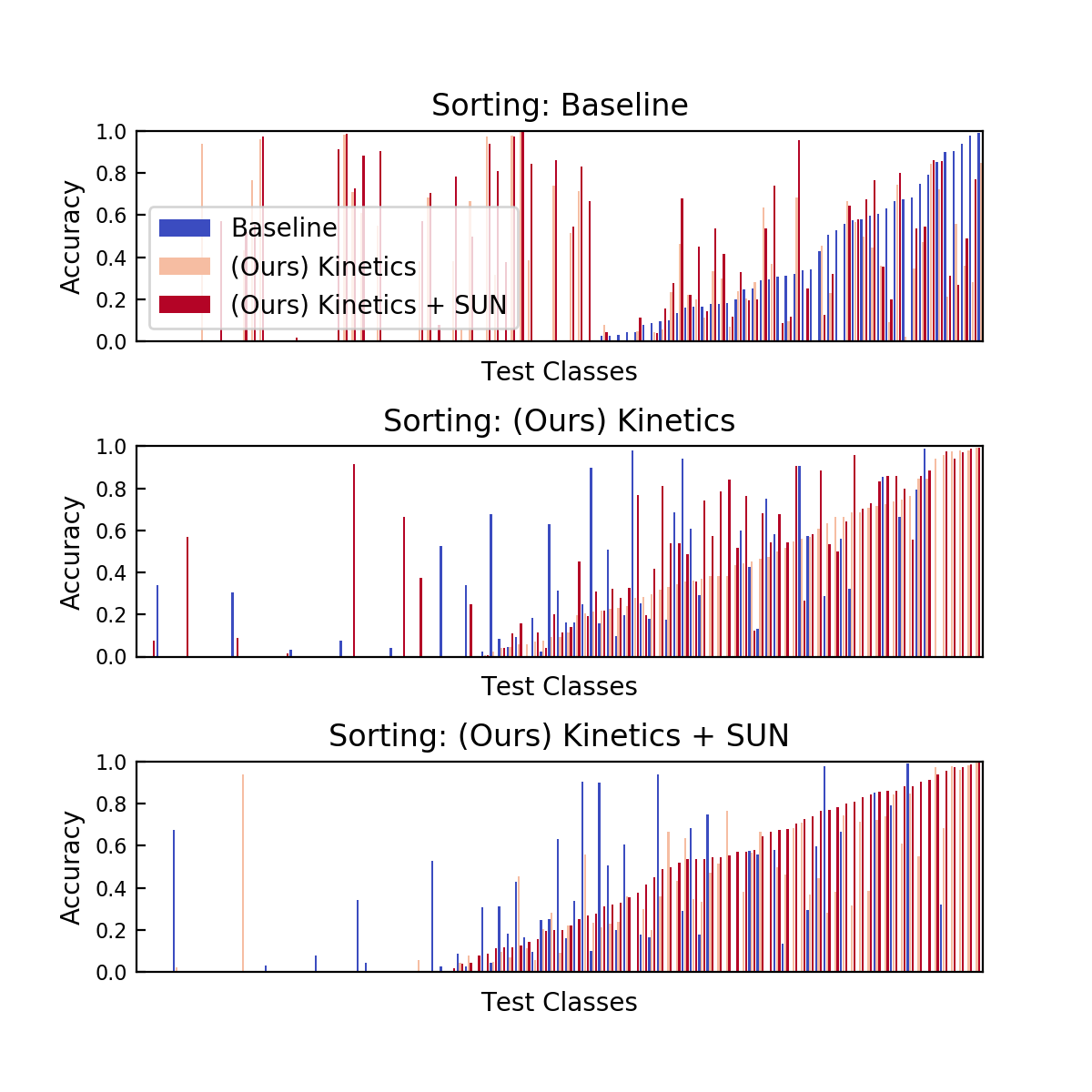}
\caption{Accuracy on each UCF101 test class, for three models. Each subplot uses different model's accuracies to sort the classes, otherwise the numbers are the same.}
\label{fig:accuracy_sorted}
\end{figure}

\begin{figure}
\centering
\includegraphics[width=.49\textwidth]{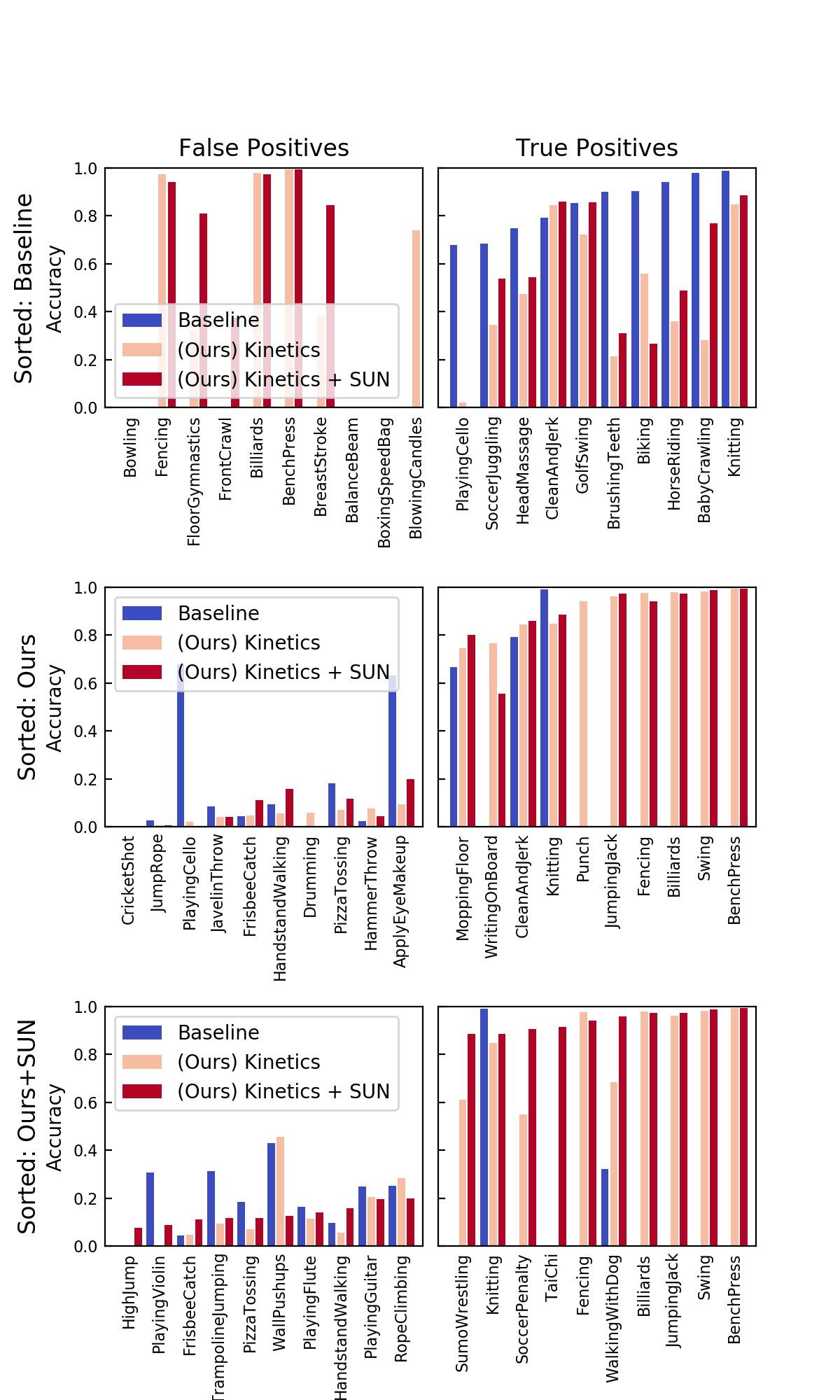}
\caption{Accuracy on best and worst 10 classes for each model.}
\label{fig:best_worst_classes}
\end{figure}

Sup.~Figure~\ref{fig:accuracy_sorted} shows the accuracy on each UCF test class for three models: baseline, e2e trained on Kinetics, and e2e pretrained on SUN397 and then trained on Kinetics. We sorted the classes from hardest to easiest for each model. Sup.~Fig.~\ref{fig:best_worst_classes} shows the same information, zoomed in on worst and best actions only. The two plots show that some of the actions which are difficult for the baseline model are correctly classified by our e2e models. On the other hand, the inverse situation is rare. In addition, the actions which are correctly classified by the baseline are also easily identified by our models.

In addition, the results of e2e trained on Kinetics and e2e pretrained on SUN and trained on Kinetics are highly correlated, but the second achieves overall better performances. This suggests that SUN provides \textit{complementary} information to Kinetics which are useful for the target task. On the other hand, the baseline is less correlated with the e2e results, suggesting that the fixed visual features have a lot to learn and \textit{should be fine-tuned}.

\subsection{Confusion matrices}
\begin{figure*}[b]
\centering
\includegraphics[width=.99\textwidth]{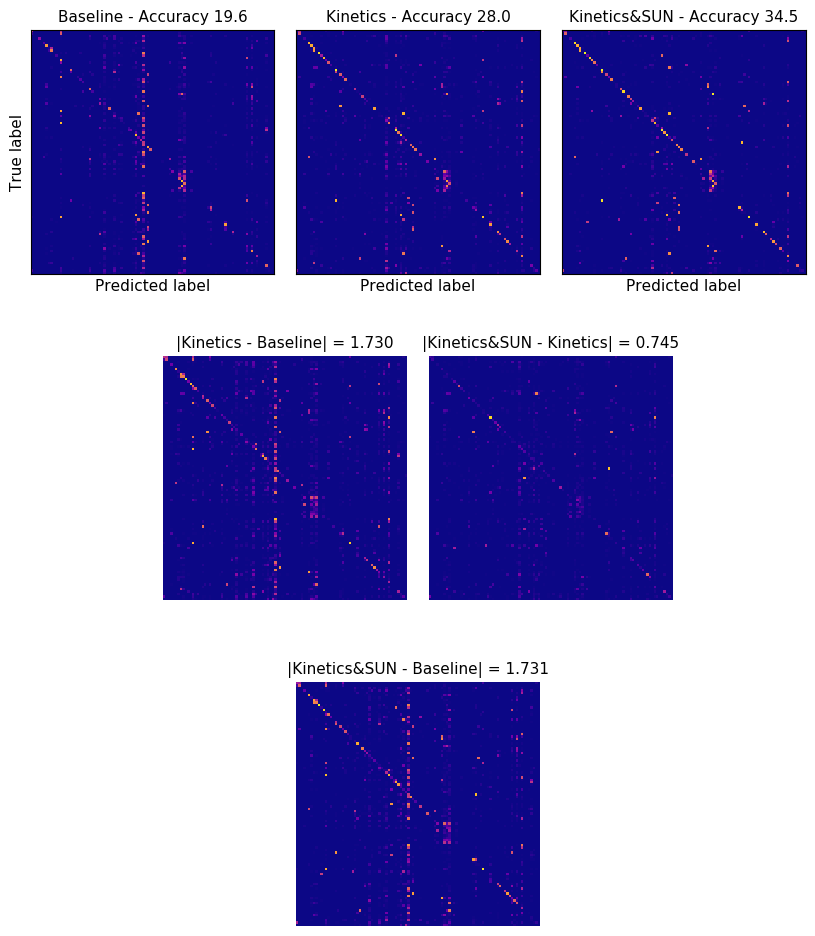}
\caption{\textbf{Top}: UCF101 confusion matrices. \textbf{Middle and Bottom}: Pairwise L2 distances between the CMs, with average score indicated in the title.}
\label{fig:cm_summary}
\end{figure*}

\begin{figure*}[b]
\centering
\includegraphics[width=.99\textwidth]{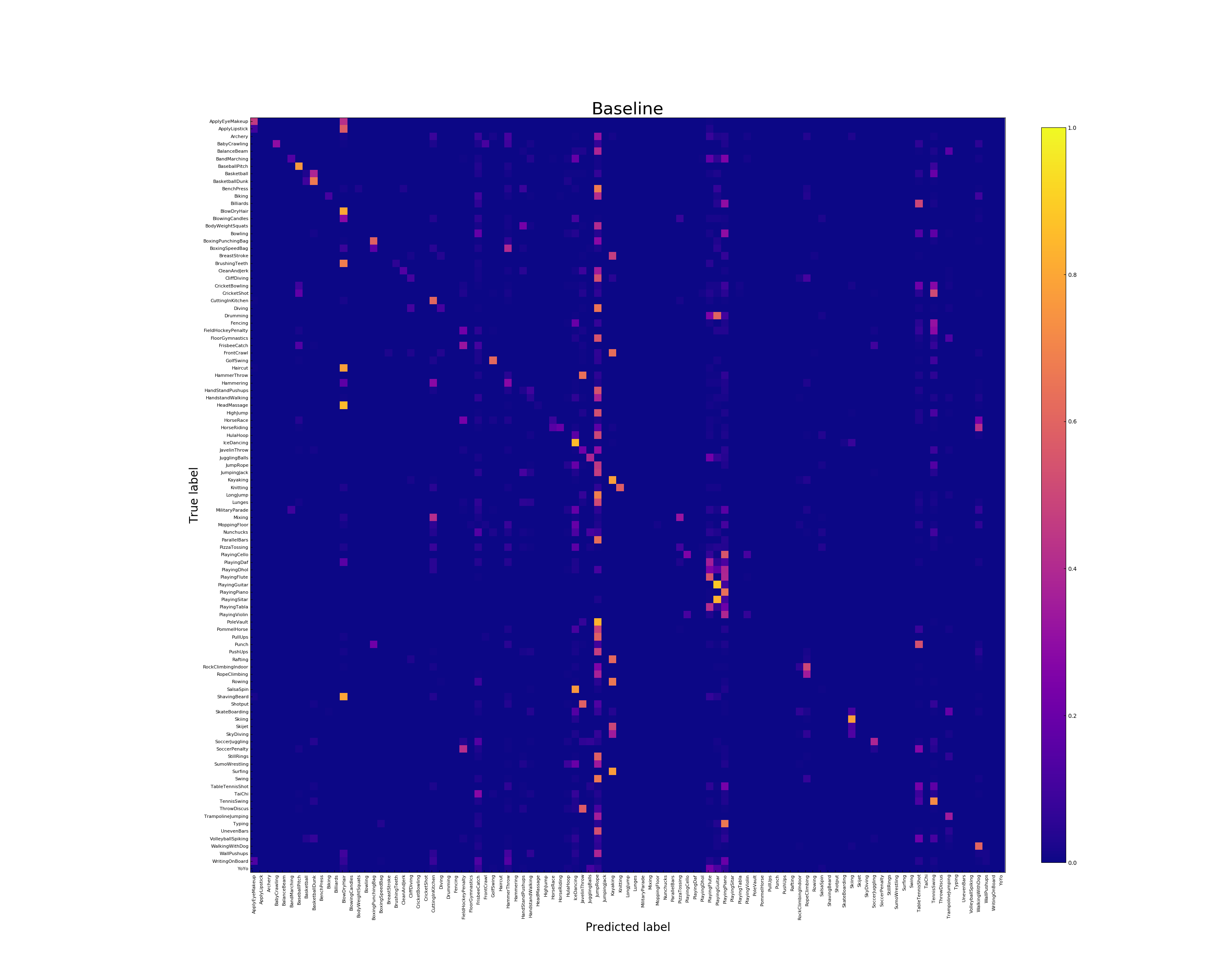}
\caption{Confusion matrix on UCF101 using our baseline model (see Sec.~3.2 in the main paper). (Figure better seen zoomed in on the digital version)}
\label{fig:cm_pretrained}
\end{figure*}

\begin{figure*}
\centering
\includegraphics[width=.99\textwidth]{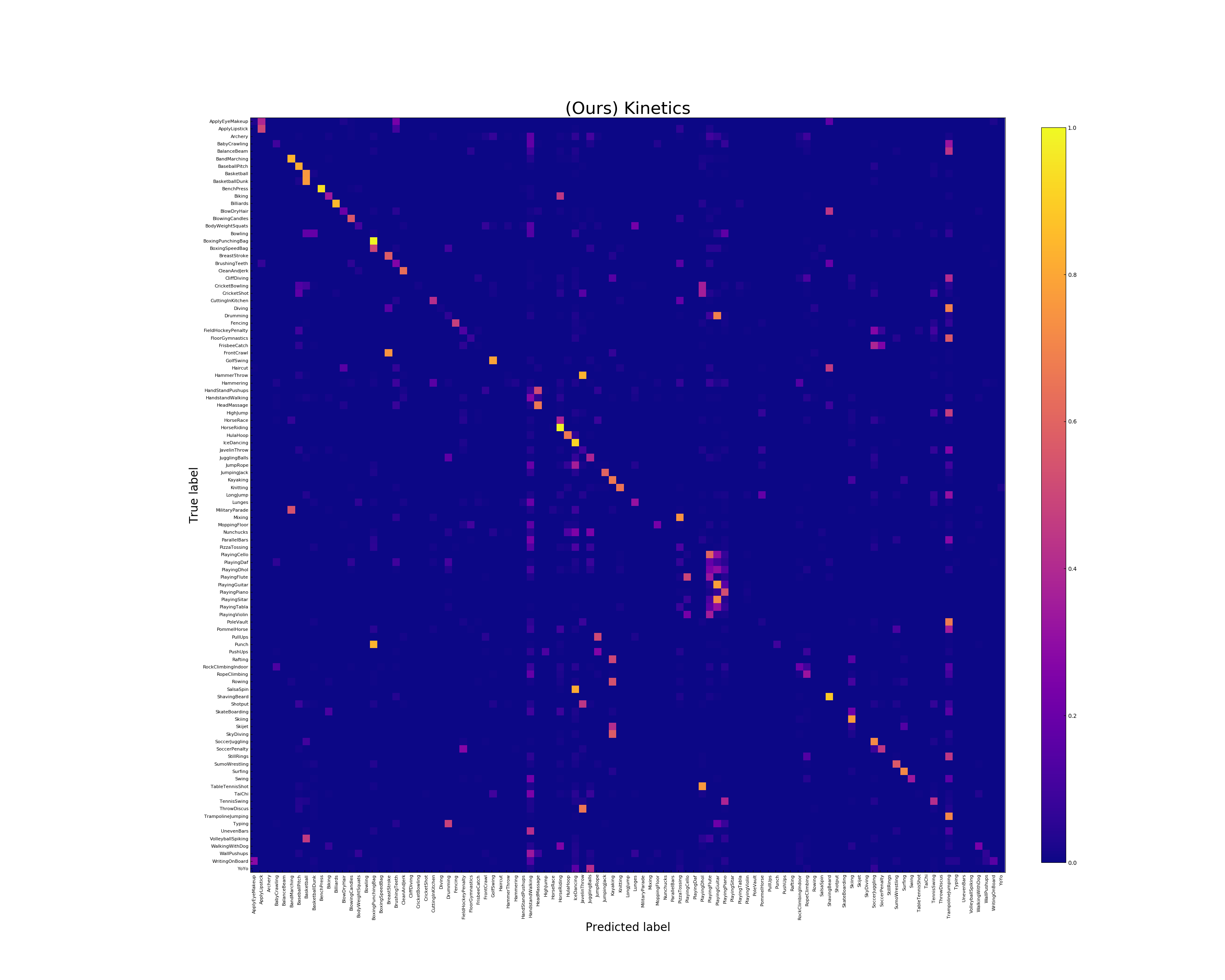}
\caption{Confusion matrix on UCF101 using our e2e model trained on Kinetics. (Figure better seen zoomed in on the digital version)}
\label{fig:cm_end2end}
\end{figure*}

\begin{figure*}
\centering
\includegraphics[width=.99\textwidth]{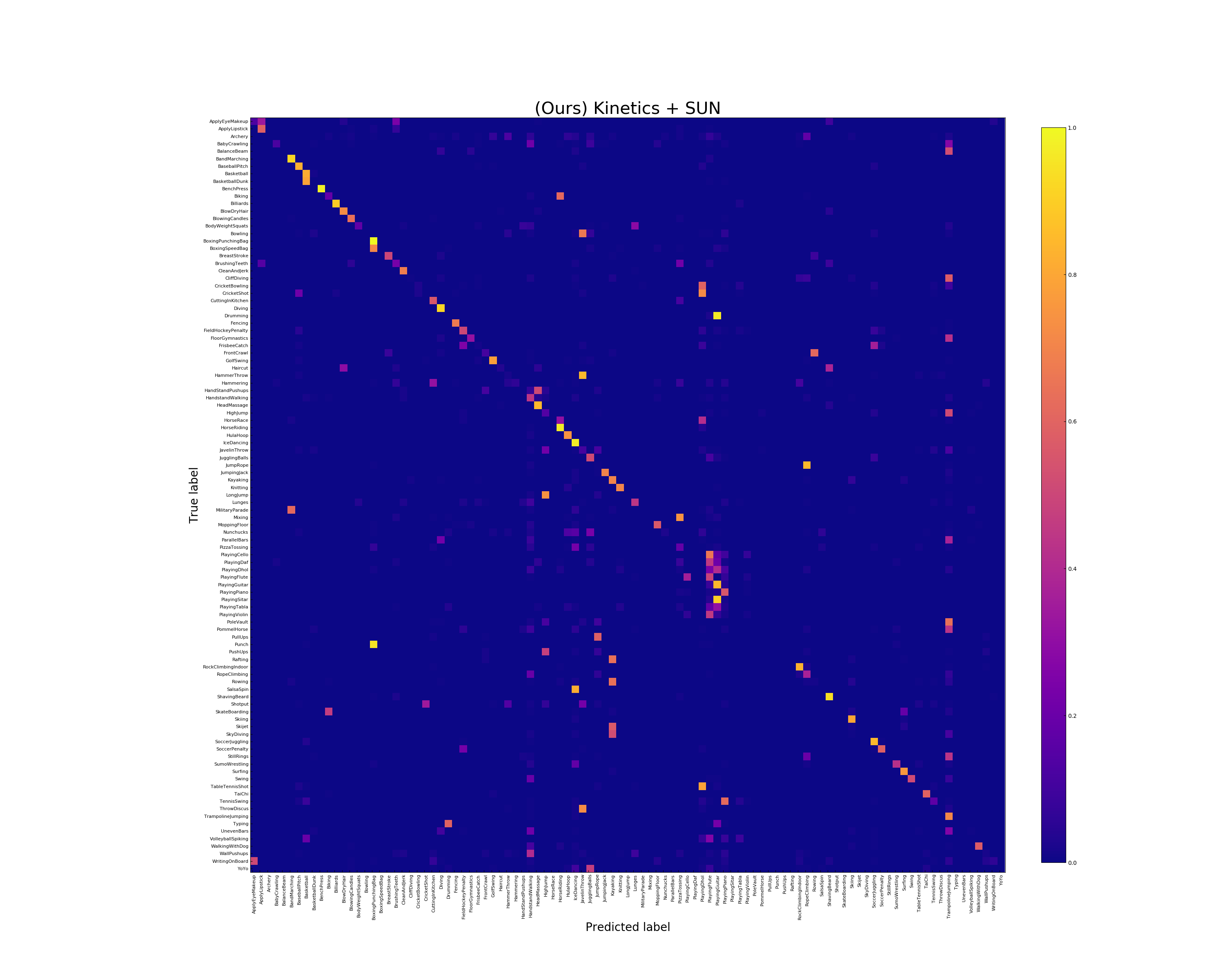}
\caption{Confusion matrix on UCF101 using our e2e model pretrained on SUN397 (see Sec.~3.4 in the main paper) and fine-tuned on Kinetics. (Figure better seen zoomed in on the digital version)}
\label{fig:cm_sun}
\end{figure*}

Sup.~Figures~\ref{fig:cm_pretrained}~-~\ref{fig:cm_sun} show confusion matrices of the three models we evaluated on UCF101. Sup.~Fig.~\ref{fig:cm_summary} shows the three CM directly compared with each other. In particular, we show the L2 distance computed pair-wise between the CMs. This shows biases present in the Baseline model, which were removed by e2e training. Some interesting biases we discovered:

\begin{compactenum}
  \item[\textbf{Playing:}] All models confuse the classes starting with the word "Playing". This issue probably comes from the way we embed the class name into the semantic embedding -- simply averaging the words. Future work might focus on tackling this problem by using a different semantic encoder. This bias is less pronounced in e2e models.
  \item[\textbf{JumpingRope:}] The baseline model wrongly classifies many actions as \emph{JumpingRope}. 
  \item[\textbf{HandStandWalking:}] Our model trained on only Kinetics has a bias towards \emph{HandStandWalking} class. This is attenuated by pre-training on SUN.
\end{compactenum}

\clearpage

\clearpage

\clearpage

\clearpage

% {\small
% \bibliographystyle{ieee_fullname}
% \bibliography{egbib}
% }